\documentclass[pdflatex,sn-mathphys]{sn-jnl}

\jyear{2021}%

\theoremstyle{thmstyleone}%
\theoremstyle{thmstyletwo}%

\theoremstyle{thmstylethree}%

\begin{document}

\title[From SMOTE to Mixup for Deep Imbalanced Classification]{From SMOTE to Mixup for Deep Imbalanced Classification}


\author[1]{\fnm{Wei-Chao} \sur{Cheng}}\email{wccheng3011@gmail.com}

\author[1]{\fnm{Tan-Ha} \sur{Mai}}\email{d10922024@csie.ntu.edu.tw}

\author*[1]{\fnm{Hsuan-Tien} \sur{Lin}}\email{htlin@csie.ntu.edu.tw}

\affil*[1]{\orgdiv{Department of Computer Science and Information Engineering}, \orgname{National Taiwan University}, \country{Taiwan}}




\abstract{Given imbalanced data, it is hard to train a good
classifier using deep learning because of the poor generalization of 
minority classes.
Traditionally, the well-known synthetic minority oversampling technique (SMOTE) 
for data augmentation, a data mining approach for imbalanced learning, has
been used to improve this generalization. However, it is unclear whether SMOTE also
benefits deep learning. In this work, we study why the original
SMOTE is insufficient for deep learning, and enhance SMOTE using
soft labels. Connecting the resulting soft SMOTE with Mixup, a modern data
augmentation technique, leads to a unified framework
that puts traditional and modern data augmentation techniques under the same
umbrella.
A careful study within this framework shows that Mixup improves 
generalization by implicitly achieving uneven margins between majority and
minority classes. We then propose a novel margin-aware Mixup technique that
more explicitly achieves uneven margins.
Extensive experimental results demonstrate that our proposed technique yields
state-of-the-art performance on deep imbalanced classification while
achieving superior performance on extremely imbalanced data. The code is open-sourced in our developed package \href{https://github.com/ntucllab/imbalanced-DL}{imbalanced-DL}} to foster future research in this direction.




\maketitle

\section{Introduction}\label{sec1}

Imbalanced classification is an old yet practical research problem for the data mining community. For example, fraud detection applications~\cite{CITEAOJ2017, CITERA2018} are often characterized by data imbalance, because there are far fewer fraudulent cases than normal ones. Another example is real-world image data for computer vision, which often exhibits long-tail properties, where minority classes occur less frequently~\cite{CITEVHG2017, CITETL2014, CITEZY2019}.

One immediate challenge in imbalanced classification is that minority classes are under-represented in the objective function, which can result in underfitting to these minority classes. This is typically addressed via re-weighting~\cite{CITEHC2019, CITEYC2019} or  resampling~\cite{CITENC2002,CITEHH2008} techniques. Re-weighting techniques belong to the family of algorithm-oriented approaches, which directly modify the objective function and optimization steps. Re-sampling techniques, on the other hand, belong to the family of data-oriented approaches, which manipulate the data being fed to the learning model.

Among algorithm-oriented techniques, re-weighting by inverse class frequencies stands out as one of the simplest methods, as discussed in previous works~\cite{CITEHC2016}. Other approaches assign weights in various ways as~\cite{CITELX2006,CITEYC2019}. For instance, in the study by Cui et al.~\cite{CITEYC2019}, a theoretical framework is developed to calculate the effective number of examples for each class, subsequently assigning suitable weights based on this calculated value.
More sophisticated approaches in the algorithm-oriented family modify the objective function to favor minority classes. For instance, the label-distribution-aware margin (LDAM) loss proposed in \cite{CITEKC2019} is based on a theoretical framework that gives minority classes a larger margin. LDAM achieves state-of-the-art
performance on benchmark datasets. Nevertheless, it is harder to optimize LDAM loss across general deep learning models due to its sophisticated design.

The most basic approaches in the data-oriented family involve oversampling
minority classes or downsampling majority
classes~\cite{CITENC2002} in an attempt to make the data distribution less
skewed. Compared with re-weighting approaches, such sampling approaches
tend to be less stable. Moreover, oversampling or downsampling
from the original data brings no new information to the learning model. 
Advanced approaches in the data-oriented family are thus based on
\emph{synthetic} (or virtual) examples, such as the well-known synthetic minority
oversampling technique (SMOTE)~\cite{CITENC2002}. As its name suggests, SMOTE
synthesizes virtual examples from minority classes to improve imbalanced
classification. The concept of SMOTE has inspired various follow-up studies that also
synthesize virtual examples for imbalanced
classification~\cite{CITEHH2005BorderSMOTE, CITECB2006SAFESMOTE, CITEHH2008,
CITEBS2019}. SMOTE and its follow-ups are closely related to data
augmentation techniques commonly used in modern deep
learning~\cite{CITEDT2017, CITEIH2018, CITEHZ2018}. Nevertheless, despite the
practical success of SMOTE for non-deep models~\cite{CITEMJ2015SVMSMOTE,
CITEBS2019}, SMOTE has not been thoroughly studied for its validity when coupled
with modern deep learning models.

A recent follow-up to SMOTE, designed for addressing imbalanced learning in the context of modern deep learning, is DeepSMOTE~\cite{DeepSMOTE2021}. This method leverages the concept of Generative Adversarial Networks (GANs)~\cite{GANs_2014} for oversampling. Effective SMOTE-based generation of synthetic examples is achieved by utilizing a deep encoder/decoder model to convert the original data into a lower-dimensional representation space. It allows DeepSMOTE to perform better on complex data than the original SMOTE.
DeepSMOTE is claimed to produce high-quality synthetic examples to assist imbalanced classification. Somehow to the best of our knowledge, DeepSMOTE needs more benchmarks to demonstrate its practical potential.


Another oversampling technique is Major-to-minor Translation (M2m)~\cite{M2mCVPR2020}. M2m addresses class imbalance by augmenting less-frequent classes through sample translation from more-frequent ones. It employs a pre-trained model to identify potential samples by introducing random noise to majority-class images. In case, the pre-trained model does not identify synthetic data, it uses existing minority samples to achieve balance. By leveraging and integrating the diversity of majority information, this approach enables the classifier to acquire more generalized features from the minority classes. Despite its benefits, M2m is computationally intensive and complex to implement due to the translation process.



In this work, we examine the SMOTE approach to
understand its disadvantages when coupled with modern deep learning models. We
correct these disadvantages via a soft variant of SMOTE that achieves
competitive performance on benchmark datasets.
We then show that the soft variant of SMOTE is coincidentally connected with
Mixup~\cite{CITEHZ2018}, a modern and popular augmentation technique for deep
learning, which however was not originally proposed for imbalanced
classification. Although a recent workshop paper~\cite{CITEHC2020} proposes a
variant that 
modifies Mixup~\cite{CITEHZ2018} to improve deep
imbalanced classification, the effectiveness and rationale of Mixup and its
variants for deep imbalanced classification have not been adequately studied, to
the best of our knowledge. 

Inspired by LDAM~\cite{CITEKC2019}, which successfully improves deep imbalanced classification with uneven margins, we study the effectiveness of Mixup via margin statistics analysis. 
We introduce a new tool called the \emph{margin gap} between the majority and minority classes. The gap is empirically demonstrated to be loosely correlated to the accuracy in 
deep imbalanced classification. We find that Mixup~\cite{CITEHZ2018} implicitly improves the margin gap, which constitutes a new piece of empirical evidence that explains its effectiveness. We further demonstrate that the gap can be more explicitly fine-tuned by making Mixup
margin-aware when synthesizing the inputs and output of the virtual example. The proposed margin-aware Mixup (MAMix) approach empirically achieves state-of-the-art performance on common imbalanced classification benchmarks, and achieves significantly better performance than Mixup and LDAM for extremely imbalanced datasets. The results validate the usefulness of our study and our proposed approach.

To make deep imbalanced learning easier for researchers and real-world users, we further develop an open-sourced python package called \textbf{imbalanced-DL} for this community. From our experience, we observed that to tackle deep imbalanced classification effectively, a single model may not be sufficient, thus we provide several strategies for people to use. The package not only implements several popular deep imbalanced learning strategies, but also provides benchmark results on several image classification tasks. We hope that our research findings along with our developed software can not only help with reproducibility but also shed lights on more comprehensive research in this community in the future.


We summarize our contributions
as the following:
\begin{enumerate}
	 \item We systematically design and
	 study the variants of the SMOTE algorithm
	 for deep learning.
	 \item We are first to utilize margin statistics to analyze whether a model
	 has learned proper representations through uneven margins for 
	 deep imbalanced classification.
	 \item We determine that a direct application of the original 
	 Mixup~\cite{CITEHZ2018} already achieves 
	 competitive
	 results for imbalanced
	 learning by implicitly enforcing uneven margins.
	 \item We further develop a simple yet effective algorithm that guides Mixup to take margins into account more explicitly, and show that the algorithm works particularly well when the data is extremely imbalanced.
\end{enumerate}

\section{Related Work}\label{secrelated}
In this section, we first define the imbalanced learning problem and review existing solutions. Then we discuss studies that are closely related to our approach. For a more comprehensive survey, see \cite{CITEJJ2019}.

\subsection{Problem Setup and Notations}
We consider the imbalanced $K$-class classification problem. Let $x \in \mathbb{R}^d$ denote the input and $y \in \{1,\dots,K\}$ denote the corresponding label. Given the training data $\mathcal{D} = \{(x_i, y_i)\}_{i=1}^n$ generated from some unknown $P(x, y)$ independently, our goal is to learn a classifier $f(x)\colon \mathbb{R}^d \to \{1,\dots, K\}$, which predicts the correct label from a given input $x$. Let $n_j$ be the size of class $j$. We assume the training data to be \emph{imbalanced}. That is, the size of the largest class $\max_i n_i$ is very different from the size of the smallest class $\min_i n_i$. The larger classes are generally called the \emph{majority}, and the smaller ones are called the \emph{minority}. After learning $f(x)$, we follow~\cite{CITEKC2019} to evaluate its accuracy on a \emph{balanced} test set generated from the same $P(x \mid y)$ for each class. The evaluation essentially equalizes the importance of each class.

In this work, we adopt two standard benchmark settings to generate controllable synthetic datasets from real-world datasets~\cite{CITEKC2019, CITEBM2018}. Both settings first decide the target size of each class by some parameters, and randomly sample within the real-world dataset to obtain the corresponding synthetic dataset under the target sizes. Both settings are based on the parameter of \emph{class imbalance ratio}, which is the ratio between the size of the largest (head) class and that of the smallest (tail) class, that is, $\rho = \max_i n_i \mathbin {/} min_i n_i$. The parameter characterizes the difficulty level of the dataset.

The first setting is called~\emph{step imbalance}, defined by $\rho$ and another parameter $\mu$. Step imbalance requires that $\mu K$ of the classes be the minority, and the other $(1-\mu)K$ be the majority. All the minority classes are of the same size, and so are all the majority classes. Following the class imbalance ratio, the size of the majority classes is $\rho$ times larger than that of the minority ones.

\begin{figure}
  \subfloat[$\rho = 100$]{
	\begin{minipage}[c][1\width]{
	   0.5\textwidth}
	   \centering
	   \includegraphics[width=1\textwidth]{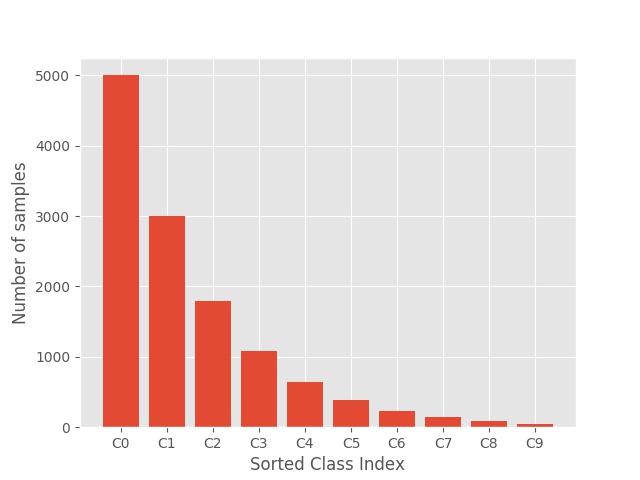}
	\end{minipage}}
  \subfloat[$\mu = 0.5,  \rho = 10$]{
	\begin{minipage}[c][1\width]{
	   0.5\textwidth}
	   \centering
	   \includegraphics[width=1\textwidth]{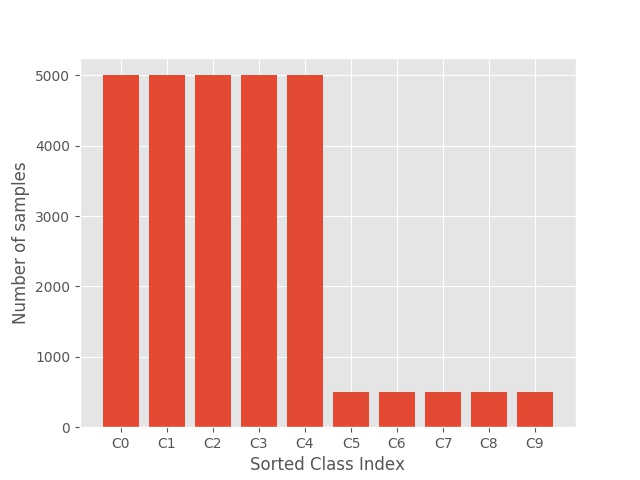}
	\end{minipage}}
\caption{Number of training samples per class in artificially created
imbalanced CIFAR-10 datasets for (a) long-tailed imbalance with $\rho=100$ and (b) step imbalance with $\rho=10, \ \mu=0.5$ }
\label{fig1}
\end{figure}

The second setting is called \emph{long-tailed imbalance}~\cite{CITEYC2019, CITEKC2019} defined by $\rho$, where the sizes of the classes follow an exponentially decreasing sequence with a decreasing constant of $\rho^{1/(K-1)}$. The constant ensures that the class imbalance ratio is exactly $\rho$. An illustrative example for long-tailed and step imbalance is in Fig.~\ref{fig1}.

\subsection{Algorithm-Oriented Approach}
Traditionally, many classification approaches are designed from the principle of empirical risk minimization (ERM), which minimizes the summation of some loss function on each example. For the imbalanced classification, the ERM principle easily leads to underfitting the minority classes, as they are under-represented in the summation.

Approaches that improve ERM for the imbalanced classification problem can be roughly divided to two categories: algorithm-oriented and data-oriented. 
One possible algorithm-oriented approach, known as cost-sensitive learning, gives a higher cost when mis-classifying the minority class~\cite{CITEKHS2018}. Cost-sensitive learning can also be carried out by giving larger weights to the minority examples. For instance, the class balance (CB) loss~\cite{CITEYC2019} re-weights each class by calculating its effective number of examples. Re-weighting increases the importance of the minority examples in the loss function, therefore preventing underfitting the minority classes.

Studies on imbalanced deep learning reveals some interesting behavior during the training process of deep learning, which generally updates the representation of the inputs and the classifier through iterative optimization.
It is shown~\cite{CITEKC2019, CITEBK2020, CITEYY2020} that learning with
re-weighting from the beginning of training can result in degraded
representations because of early overfitting to the minority classes, making the performance of the re-weighting even worse than ERM. To solve the overfitting issue,  \cite{CITEKC2019} proposed the
deferred re-weighting (DRW) technique.
DRW splits the one-stage training of deep learning into two phases. In the first phase, ERM without any re-weighting is used to learn a good representation, with the hope of not overfitting to the minority classes. Then, the training continues with an annealed (smaller) learning rate on a re-weighted loss, such as CB loss, in the second phase.

With the DRW technique, some other algorithmic attempts are used to improve ERM. Label-distribution-aware margin (LDAM)~\cite{CITEKC2019} follows the rich literature of margin classifiers~\cite{CITEWL2016, CITEWF2018}
and proposes a loss function that encourages class-dependent margins to tackle the class imbalance issue. The ideal margin $\tau_i$ for each class is theoretically derived to be proportional to $n_i^{1/4}$. That is,
\begin{equation}
    \tau_i = \dfrac{C}{n_i^{1/4}}\label{eq1}
\end{equation}
with some constant $C$. The ideal margin hints the need to enforce larger margins for the minority classes. 

With the definition of $\tau_i$, the authors of LDAM propose a margin-aware loss function that can be used in both the ERM phase and the re-weighting phase of DRW. Combining LDAM and DRW with the CB loss in the second phase results in a state-of-the-art approach for imbalanced learning~\cite{CITEKC2019}, which will serve as the baseline of our comparison.

\subsection{Data-Oriented Approach}
A common approach for imbalanced multi-class classification at the data level
is undersampling for majority classes or oversampling for minority classes. 
One such approach is SMOTE~\cite{CITENC2002}, which essentially oversamples
minority classes by creating artificial examples through k-nearest
neighbors within the same class. In the context of deep learning, this 
kind of oversampling can be viewed as a type of data augmentation.
Also note ADASYN~\cite{CITEHH2008} and LoRAS~\cite{CITEBS2019}, 
SMOTE extensions that address class imbalance using machine
learning approaches. In this work, we revisit SMOTE
and incorporate it into a modern deep learning pipeline.

\subsubsection{SMOTE}
Traditional replication-based oversampling techniques are prone to
overfitting. To account for this, \cite{CITENC2002} propose
oversampling by creating synthetic examples for minority
classes; in this case, the synthetic examples are thus not replicated.
Specifically, for those samples categorized as belonging to a minority class,
they create new data points by interpolating them with their 
k-nearest neighbors which belong to the same categories. Note that at the
time this technique was proposed, deep learning techniques were not yet widely
used. Thus, we first study this technique and design two SMOTE-like
techniques along with the current end-to-end deep learning training pipeline.
This is described in detail in the next section. We also note DeepSMOTE~\cite{CITEDEEPSMOTE}, which was published during the course of the current study. However, since this
approach requires two-stage training in which the first stage requires 
training an encoder-decoder framework, followed by DeepSMOTE generation, we
consider it to be aligned more with GAN-based work, which is not our main focus.

\subsection{Mixup-based Techniques}
\subsubsection{Mixup}
One of the most famous regularization---or data augmentation---techniques in deep
neural networks for image classification problem is Mixup~\cite{CITEHZ2018},
which constructs virtual training examples via simple linear combinations
as 
\begin{equation}
    \Tilde{x} = \lambda x_i + (1 - \lambda) x_j\label{eq2}
\end{equation}
\begin{equation}
    \Tilde{y} = \lambda y_i + (1 - \lambda) y_j, \label{eq3}
\end{equation}
in which $(x_i, y_i)$ and $(x_j, y_j)$ are two examples drawn uniformly from the
training data and $\lambda \in [0, 1)$. Mixup-based techniques have been shown
to mitigate the memorization of corrupt labels, increase 
robustness to adversarial training, and improve the generalizability of
deep networks, which has led to state-of-the-art performance on tasks such as image
classification. 

\subsubsection{Remix}
Similar to our proposed method, a recently proposed technique called Remix~\cite{CITEHC2020} relabels Mixup-created examples with minority class labels chosen by two hyperparameters in $\tau$ and $P$-majority. Specifically, Remix can be formulated as 
\begin{equation}
    \Tilde{x} = \lambda_x x_i + (1 - \lambda_x) x_j\label{eq4}
\end{equation}
\begin{equation}
    \Tilde{y} = \lambda_y y_i + (1 - \lambda_y) y_j. \label{eq5}
\end{equation}
Here, $(x_i, y_i)$ and $(x_j, y_j)$ are two examples drawn at random from the
training data and $\lambda_x \in [0, 1)$, where it is sampled from the beta
distribution as introduced by the original Mixup authors~\cite{CITEHZ2018}. Note
that in contrast to the original Mixup, where the mixing
factor $\lambda$ is the same for both $x$ and $y$, in Remix, this criterion
is relaxed; they relabel the mixing factor for the label
$\lambda_y$ according to the following conditions:
\begin{equation}
    \lambda_y = \left\{
        \begin{aligned}
            0 & , & {n_i} \mathbin{/} {n_j} \geq P \ \text{and} \ \lambda_x < \tau \\
            1 & , & {n_i} \mathbin{/} {n_j} \leq {1}  \mathbin{/} {P} \ \text{and} \ 1-\lambda_x < \tau \\
            \lambda_x & , & \text{otherwise} \\
        \end{aligned}
    \right. ,\label{eq6}
\end{equation}
where \cite{CITEHC2020} defines $P$ as P-majority. 
Specifically, this is given a mixup-created sample pair $(x_i, y_i)$ and $(x_j, y_j)$,
where the respective sample numbers for classes $i$ and~$j$ are denoted as
$n_i$ and $n_j$. For this
example pair, if ${n_i} \mathbin{/} {n_j} \geq P$, then $x_i$ is defined to
be \emph{P-majority} over $x_j$. Thus the idea of Remix
is to favor minority classes by examining each selected Mixup-selected pair: 
if one is P-majority over the other, and the other hyperparameter criterion is
also met, than Remix relabels the synthesized example with the label of the
minority class. Note that here it is suggested that $\tau$ be set to 0.5 and $P$ be set
to 3~\cite{CITEHC2020}. 

\section{Main Approach}

\begin{figure}
    \centering
    \includegraphics[width=.8\linewidth]{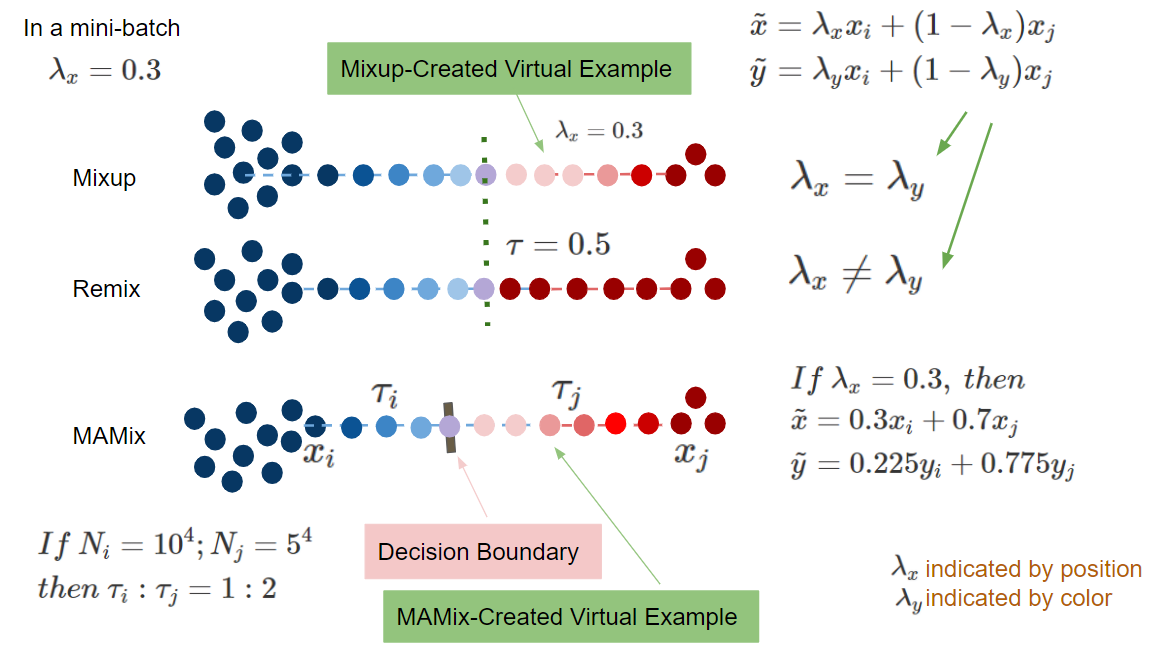}
	 \caption{Mixup Framework Illustration}
    \label{figillustration}
\end{figure}

We observe that Mixup~\cite{CITEHZ2018} and Remix~\cite{CITEHC2020} can generalize to a general framework, in the sense that they both train with similar fashion. We term this~\emph{Mixup framework} (Fig.~\ref{figillustration}), and describe the training algorithm for Mixup framework in Algorithm~\ref{algo1}.

\begin{algorithm}
\SetAlgoLined
\textbf{Required} {Dataset $D = \{(x_i, y_i)\}_{i=1}^n$, model with parameter $\theta$}\\
 Initialize\;
 \While{training}{
  Sample  $\{(x_i, y_i), (x_j, y_j)\}_{m=1}^M$ from D\;
  Sample $\lambda_x$ $\sim$ $\mathit{Beta}(\alpha, \alpha)$\;
  \For{m = 1 to M}{
    (a) Obtain mixed input $\Tilde{x}$ \;
    (b) Obtain $\lambda_y$ \;
    (c) Obtain mixed label $\Tilde{y}$ \;
  }
  $\mathcal{L}(\theta)$ $\gets$ $\dfrac{1}{M} \sum_{(\Tilde{x}, \Tilde{y})} L((\Tilde{x}, \Tilde{y});\theta)$\;
  $\theta$ $\gets$ $\theta - \delta\bigtriangledown_{\theta} \mathcal{L}(\theta)$;
}
\caption{Mixup Framework Training Algorithm}
\label{algo1}
\end{algorithm}

Specifically, within this Mixup Framework, the main difference between each method lies in three steps during mini-batch training, that is, (a) How to obtain mixed input (b) How to obtain label mixing factor $\lambda_y$ and (c) How to obtain mixed label. For example, Mixup~\cite{CITEHZ2018} obtains mixed input through~\eqref{eq2}, mixed label through~\eqref{eq3}, and its $\lambda_y = \lambda_x$; while Remix~\cite{CITEHC2020} obtains mixed input by~\eqref{eq4}, mixed label by~\eqref{eq5}, and $\lambda_y$ by~\eqref{eq6}.

With this Mixup Framework, we design new methods through two perspectives. First, we design two SMOTE-like techniques---SMOTE-Mix and Neighbor-Mix---within this framework to examine the effectiveness of SMOTE in modern deep learning from input mixing perspective, and this is described in the following~\emph{Approach 1}. Secondly, we propose to incorporate the idea of~\emph{uneven margin} into this Mixup framework to better tackle deep imbalanced learning, which will be illustrated in~\emph{Approach 2}. Our proposed Approach 2 and Remix can be viewed from non-uniform label mixing perspective.

\subsection{Approach 1: SMOTE-like Techniques}
We introduce two SMOTE-like techniques from input mixing perspective in SMOTE-Mix and Neighbor-Mix. First, we perform SMOTE-like input mixing under Mixup framework and term this \emph{SMOTE-Mix}. Recall that SMOTE performs linear interpolation with their
same-class samples on input only. Formally, with SMOTE-Mix, we 
create synthetic examples from two training samples $(x_i, y_i), (x_j, y_j)$
with the following equations:
\begin{equation}
    \Tilde{x} = \lambda x_i + (1 - \lambda) x_j\label{eq10}
\end{equation}
\begin{equation*}
    x_j = \text{same-class nearest neighbor of } x_i
\end{equation*}
\begin{equation}
    \Tilde{y} = y_i. \label{eq11}
\end{equation}
Following Algorithm~\ref{algo1}, SMOTE-Mix obtains mixed input by~\eqref{eq10}, mixed label by~\eqref{eq11}, and $\lambda_y = \lambda_x$.
Note that in SMOTE-Mix, the mix pair for creating synthetic examples is
sampled from its same-class nearest neighbors. Thus for each pair, the
label is the same ($y_i = y_j$); that is, they are hard labels.

We then further relax the above idea by not restricting $x_j$ to be the same
class as $x_i$; that is, we still create synthetic samples through the nearest
neighbors, but due to the fact that data are in a high dimensional space, its
nearest neighbors may not belong to the same categories. We term this relaxed
version \emph{Neighbor-Mix}, and formulate it as 
\begin{equation}
    \Tilde{x} = \lambda x_i + (1 - \lambda) x_j \label{eq12}
\end{equation}
\begin{equation*}
    x_j = \text{nearest neighbor of } x_i
\end{equation*}
\begin{equation}
    \Tilde{y} = \lambda y_i + (1 - \lambda) y_{x_j}. \label{eq13}
\end{equation}
Following Algorithm~\ref{algo1}, Neighbor-Mix obtains mixed input by~\eqref{eq12}, mixed label by~\eqref{eq13}, and $\lambda_y = \lambda_x$.
Note that for $\Tilde{y}$, Neighbor-Mix is soft-label, as $x_j$ may belong to other categories.

We discuss the empirical results of SMOTE-Mix and Neighbor-Mix on modern
long-tailed image datasets in Table~\ref{tab1} to verify the effectiveness of SMOTE
in deep learning. Now we further propose our main strategy within the Mixup
framework to address deep imbalanced classification.

\subsection{Approach 2: Margin-Aware Mixup (MAMix)}
Inspired by the attempt to achieve uneven margins
through a well-designed LDAM loss~\cite{CITEKC2019}, we propose incorporating the concept of
uneven margins into Mixup-based data augmentation techniques. We adopt the common definition and define the margin of an example $(x, y)$ as
\begin{equation}
    \gamma(x,y) = f(x)_y - \max_{j \ne y}f(x)_j.\label{eq7}
\end{equation}
The margin for class~$j$ is defined as the average margin of all examples in the class:
\begin{equation}
    \overline{\gamma_j} = \dfrac{1}{n_j} \sum_{i: y_i=j} \gamma(x_j, y_j),\label{eq8}
\end{equation}

Recall that the optimal class-distribution-aware margin trade-off follows \eqref{eq1}~\cite{CITEKC2019}. Suppose that $(x_i, y_i)$ and $(x_j, y_j)$ are two samples of different classes. Define $\eta_i$ as the distance from $x_i$ to the decision boundary
between class~$i$ and~$j$, and define $\eta_j$ similarly. Motivated by
\eqref{eq1}, we set
\begin{equation}
    \eta_i = {1} \mathbin{/} {n_i^{\omega}}; \eta_j = {1}  \mathbin{/}  {n_j^{\omega}}.\label{eq14}
\end{equation}
We tune the hyper-parameter $\omega$ to strike the best trade-off in the proposed
margin-aware Mixup. The sensitivity of this hyper-parameter is discussed
in Table~\ref{tab10}.

The proposed margin-aware Mixup (MAMix) is formulated as 
\begin{equation}
    \Tilde{x}^{\mathit{MAM}} = \lambda_x x_i + (1- \lambda_x) x_j \label{eq15}
\end{equation}
\begin{equation}
    \Tilde{y}^{\mathit{MAM}} = \lambda_y y_i + (1 - \lambda_y) y_j. \label{eq16}
\end{equation}
Note that here, $\lambda_x$ and the Mixup-selected pair $(x_i, y_i)$
and $(x_j, y_j)$ are obtained as in the original Mixup, whereas
we compute $\lambda_y$ for each Mixup-selected pair based on the following
formula, where $\lambda_y \in [0, 1]$:
\begin{equation}
    \lambda_y = \left\{
        \begin{aligned}
            1 - \dfrac{(1 - \lambda_x) \times 0.5}{\eta_i \mathbin{/} (\eta_i+\eta_j)} & \ , & \text{if }  \lambda_x \geq \eta_j \mathbin{/} (\eta_i + \eta_j) \\
            \dfrac{(0.5) \times (\lambda_x)}{\eta_j \mathbin{/} (\eta_i+\eta_j)}  &  \ , & \text{if }  \lambda_x < \eta_j \mathbin{/} (\eta_i + \eta_j).\\
        \end{aligned}
    \right.\label{eq17}
\end{equation}
Therefore, with Algorithm~\ref{algo1}, our proposed MAMix obtains mixed input by~\eqref{eq15}, mixed label by~\eqref{eq16}, and $\lambda_y$ through~\eqref{eq17}.
Essentially, we obtain the optimal mixing factor by ${\eta_j} \mathbin{/}
{(\eta_i + \eta_j)}$; note that $\eta_i$, $\eta_j$ are obtained via
\eqref{eq14}. If the mixing factor $\lambda_x$ is exactly the same as ${\eta_j}
\mathbin{/} {(\eta_i + \eta_j)}$, the probability to output this synthetic
example should be exactly $50\%$ for class~$i$ and $50\%$ for class~$j$. Also, if
the mixing factor $\lambda_x$ is larger or smaller than ${\eta_j} \mathbin{/}
{(\eta_i + \eta_j)}$, we compute its corresponding $\lambda_y$ using
\eqref{eq17}, where the core idea is to use arithmetic progression to ensure
that the classifier favors minority classes and therefore achieves uneven
margins.

Recall that in original Mixup, the mixing factor $\lambda_x$
is the same for synthetic $x$ and $y$, that is, $\lambda_x = \lambda_y$. The
core idea for Mixup to better account for class-imbalanced learning is by making $y$~\emph{not uniform}. Our proposed method achieves this by incorporating
margin-aware concepts. We also note the recently proposed
Remix~\cite{CITEHC2020}, which is similar to our approach: 
  given two hyperparameters, they relabel the synthetic $y$ to be the minority class. An illustration is in Fig.~\ref{figillustration}.

\section{Experiments}
\subsection{Experiment Setup}
We follow \cite{CITEKC2019} in creating synthetic datasets for CIFAR-10~\cite{CITEAK2009}, CIFAR-100, and Tiny ImageNet; additionally, we follow \cite{CITEHC2020} for CINIC-10 and
\cite{CITEYY2020} for SVHN for a more complete study. Furthermore, we examine CIFAR-10 and CINIC-10 with extreme imbalance ratios to simulate extremely imbalanced scenarios. Moreover, for step imbalance, we follow~\cite{CITEKC2019} to fix $\mu = 0.5$. For a more comprehensive description about the dataset preparation, please refer to the appendix.

\comments{
\paragraph{Imbalanced CIFAR}
There were 50,000 training images and 10,000 validation images of size
32$\times$32 for both original CIFAR-10 and CIFAR-100, with 10 and 100 classes,
respectively. We created imbalanced training images with imbalance ratios of 10,
50, and 100 for general imbalanced cases, as well as imbalance ratios of 200, 250, and 300 for further evaluation on extremely imbalanced scenarios, and kept the validation images intact for both CIFAR-10 and
CIFAR-100.

\paragraph{Imbalanced CINIC-10}
We use the training and validation sets with 90,000 images each for training and evaluation. Other than the commonly used imbalance ratios of 10, 50, and 100, we also created an artificial imbalanced CINIC-10 with extreme imbalance ratios of 200, 250, and 300 for further evaluation.

\paragraph{Imbalanced SVHN}
The Street View House Numbers (SVHN) dataset containes 73,257 digits for training, and 26,032 digits for testing. We evaluate with imbalance ratios of 10, 50, and 100 as in~\cite{CITEYY2020}.

\paragraph{Imbalanced Tiny ImageNet}
The Tiny ImageNet dataset contains 200 classes, each class of which contains 500 images for training and 50 for validation. We evaluated with imbalance ratios of 10 and 100 as~\cite{CITEKC2019}.
}

\subsection{Compared methods}
We compared our method with the baseline training methods: 
(1)~Empirical risk minimization (ERM) loss, where we use standard cross-entropy
loss with all examples sharing the same weights. (2)~Deferred re-weighting
(DRW), proposed by \cite{CITEKC2019}, where we train with
standard ERM in the first stage and then apply re-weighting in
the second stage with the final learning rate decay. 
(3)~The margin-based state-of-the-art work of LDAM-DRW~\cite{CITEKC2019}.
(4)~The recent Mixup-based Remix~\cite{CITEHC2020}. 
Note that following the
notation of \cite{CITEKC2019}, when two methods are combined, we 
abbreviate their acronyms with a dash. Our main
proposed method is margin-aware Mixup (MAMix). For all experiments, we report
the mean and standard deviation over 5 runs with different seeds. We computed
the margin gap $\gamma_{\mathit{gap}}$ (introduced later) on the validation sets. Our proposed method was
developed using PyTorch~\cite{CITEPA2017}.

\section{Results and Analysis}
In this section, we first discuss SMOTE-like techniques---SMOTE-Mix and
Neighbor-Mix---for imbalanced deep learning. Then we discuss Mixup-based
approaches~\cite{CITEHZ2018}, \cite{CITEHC2020} and their effects on margin
statistics compared with margin-based state-of-the-art work in 
LDAM~\cite{CITEKC2019}.

\subsection{From SMOTE to Mixup}
When directly using SMOTE for oversampling, the performance gain from around 71\% to 72\% is not
competitive enough (Table~\ref{tab1}). Previous studies~\cite{CITEKC2019,
CITEBK2020, CITEHY2020} show that training with re-weighting or
re-sampling based approaches is harmful for representation learning with deep
models. Therefore, direct incorporation of SMOTE into deep
learning achieves only limited performance improvements. However, SMOTE-Mix and
Neighbor-Mix are effective when coupled with DRW (Table~\ref{tab1}). Neighbor-Mix coupled with DRW achieves a greater performance improvement over SMOTE-Mix, whereas the performance of Neighbor-Mix is still inferior to that of Mixup, as demonstrated in Table~\ref{tab1}, in which the
performance difference lies in how to select the Mixup pair during training.

\begin{table}
\caption{Top-1 validation accuracy (mean $\pm$ std) on long-tailed
imbalanced CIFAR-10 with ratio $\rho$ = 100 with ResNet32 using SMOTE and its
two variants}
\begin{center}
\begin{tabular}{@{}ccccccccccc@{}}
\toprule
{Method} & Accuracy \\
\midrule
ERM & 71.23 $\pm$ 0.51  \\
SMOTE & 72.68 $\pm$ 1.41 \\
DRW & 75.08 $\pm$ 0.61 \\
M2m & 76.15 $\pm$ 0.72  \\
DeepSMOTE & 76.66 $\pm$ 0.57  \\
SMOTE-Mix--DRW & 77.46 $\pm$ 0.64 \\
Neighbor-Mix--DRW & 80.44 $\pm$ 0.32  \\
Mixup--DRW & 82.11 $\pm$ 0.57  \\

\bottomrule
\end{tabular}
\label{tab1}
\end{center}
\end{table}

Motivated by the competitive results of SMOTE-Mix and Neighbor-Mix, we further
relaxed Neighbor-Mix back to the original form of Mixup to
examine the effectiveness of this approach on imbalanced data. Mixup
is a modern data augmentation technique that is widely
recognized to be effective in the deep image classification literature. However,
the datasets are usually balanced; the effect of Mixup for imbalanced
datasets has not been widely studied. Therefore, by simply applying Mixup
on imbalanced learning settings, we expect to see 
improvement over a non-Mixup counterpart. For example, in long-tailed imbalanced
CIFAR-10 with an imbalance ratio of $\rho$ = 100, we can see that the top-1
validation accuracy improves from 72\% to around 74\% (Table~\ref{tab7}) when
applying Mixup, which is expected. However, when Mixup is deployed with
DRW, the performance boosts from 72\% to around 82\% (Table~\ref{tab7}) under
the same setting, which exceeds the previous state-of-the-art
result on imbalanced learning of LDAM-DRW~\cite{CITEKC2019}. The
comprehensive results for imbalanced CIFAR-10 and CIFAR-100 are given in
Tables~\ref{tab7} and~\ref{tab8}; those for imbalanced CINIC-10
are given in Table~\ref{tab9}. The detailed results for imbalanced SVHN
and imbalanced Tiny-ImageNet are shown in Tables~\ref{tab11} and~\ref{tab12},
respectively.

\begin{table}
\caption{Top-1 validation accuracy (mean $\pm$ std) on extremely long-tailed
imbalanced CIFAR-10 using ResNet32}
\begin{center}
\begin{tabular}{@{}cccccccccccc@{}}
\toprule
{Imbalance ratio} & $200$ &  $250$  & $300$ \\ 
\midrule
{Mixup--DRW} & $77.02\pm0.53$ & $76.33\pm0.78$ & $73.39\pm0.47$ \\
{Remix--DRW} & $77.23\pm0.61$ & $75.39\pm0.72$ & $73.79\pm0.29$ \\
{MAMix--DRW} & \textbf{78.08 $\pm$ 0.23} & \textbf{76.34 $\pm$ 0.71} & \textbf{74.85 $\pm$ 0.29} \\
{MAMix-Remix--DRW} & $78.01\pm0.23$ & $76.25\pm0.63$ & $74.87\pm0.56$\\
\bottomrule
\end{tabular}
\label{tabextremecifar10}
\end{center}
\end{table}

\begin{table}
\caption{Top-1 validation accuracy (mean $\pm$ std) on extremely long-tailed
imbalanced CINIC-10 using ResNet18}
\begin{center}
\begin{tabular}{@{}cccccccccccc@{}}
\toprule
{Imbalance ratio} & $200$ &  $250$  & $300$ \\ 
\midrule
{Mixup--DRW} & $66.86\pm0.50$ & $65.24\pm0.50$ & $63.91\pm0.39$ \\
{Remix--DRW} & $66.46\pm0.51$ & $64.76\pm0.47$ & $63.25\pm0.17$ \\
{MAMix--DRW} & \textbf{67.59 $\pm$ 0.37} & \textbf{66.54 $\pm$ 0.36} & \textbf{65.27 $\pm$ 0.37} \\
\bottomrule
\end{tabular}
\label{tab2}
\end{center}
\end{table}

\begin{table}
\caption{Top-1 validation accuracy (mean $\pm$ std) on extremely imbalanced
CINIC-10 using ResNet18}
\begin{center}
\begin{tabular}{@{}cccccccccccc@{}}
\toprule
{Dataset}& \multicolumn{1}{c}{Long-tailed}  & \phantom{a} &  \multicolumn{1}{c}{Step} \\
\cmidrule{2-4}
{Imbalance ratio} & $200$ &&  $200$  \\ 
\midrule
{ERM} & 56.22 $\pm$ 1.46 &&  52.01 $\pm$ 0.52\\
{DRW} & 58.97 $\pm$ 0.30  &&  57.87 $\pm$ 1.01 \\
{LDAM--DRW} & 63.09 $\pm$ 0.54  &&  65.47 $\pm$ 0.63 \\
{Mixup--DRW} & 66.86 $\pm$ 0.50 &&  65.61 $\pm$ 0.59\\
{Remix--DRW} & 66.46 $\pm$ 0.51  &&  66.61 $\pm$ 0.27 \\
{MAMix--DRW} & \textbf{67.59 $\pm$ 0.37}  &&  \textbf{67.34 $\pm$ 0.32} \\
\bottomrule
\end{tabular}
\label{tab3}
\end{center}
\end{table}

\begin{table}
\caption{Margin gap on imbalanced CIFAR-10 with $\rho$ = 100 using ResNet32}
\begin{center}
\begin{tabular}{@{}cccccccccccc@{}}
\toprule
{Dataset}& \multicolumn{1}{c}{Long-tailed}  & \phantom{a} & \multicolumn{1}{c}{Step} & \phantom{a}\\
\cmidrule{2-4} 
{Imbalance ratio} & $100$ &&  $100$  \\ \midrule
{ERM} & 7.645  &&  8.515 \\
{DRW} & 6.089  &&  7.086 \\
{LDAM--DRW} & 0.171  &&  0.056 \\
{Mixup--DRW} & -0.978  &&  -0.481 \\
{Remix--DRW} & -1.598 &&  -1.870 \\
{MAMix--DRW} & -1.136  &&  -1.798 \\
\bottomrule
\end{tabular}
\label{tab4}
\end{center}
\end{table}

\begin{table}
\caption{Margin gap for extremely imbalanced CIFAR-10 with $\rho$ = 300 using
ResNet32}
\begin{center}
\begin{tabular}{@{}cccccccccccc@{}}
\toprule
{Method} & Margin gap \\
\midrule
Remix--DRW & -0.101  \\
MAMix--DRW & -0.487 \\
\bottomrule
\end{tabular}
\label{tab5}
\end{center}
\end{table}

Note that Mixup-based methods work best when coupled with DRW. 
Traditional re-weighting or re-sampling approaches have been shown to harm feature extraction
when learning with imbalanced data~\cite{CITEKC2019, CITEHY2020}. As a result, DRW provides a training scheme
which first learns a good representation and further accounts for 
minority classes by re-weighting at later training stages.

In general imbalanced settings where the imbalance ratios are not extreme (e.g.,
$\rho < 200$), the original Mixup coupled with DRW already
achieves competitive results, with the results among different Mixup-based
approaches comparable to each other. However, our proposed MAMix
outperforms the original Mixup and Remix in extremely imbalanced cases 
(e.g., $\rho \geq 200$), as demonstrated in Tables~\ref{tabextremecifar10}, \ref{tab2}, 
and~\ref{tab3}. When the imbalance ratio is extreme, our method consistently 
achieves results superior to those of
Mixup and Remix, demonstrating the effectiveness of our method as well as the
necessity of our algorithm in extremely imbalanced scenarios. Moreover, MAMix
also serves as a general technique used to improve over Mixup or Remix;
when deploying MAMix on top of Remix (MAMix--Remix--DRW
in Table~\ref{tabextremecifar10}), there is also improvement
(Table~\ref{tabextremecifar10}). However, simple deployment of MAMix already
yields superior results.

To further demonstrate the effectiveness of our proposed method, we can see from Table~\ref{tab15} for detailed per class accuracy evaluation. As we can see from Table~\ref{tab15}, with ERM, the minority classes (i.e, C7,C8,C9), the accuracy for those classes are low, with C8 and C9 to be 0.46 and 0.48 respectively. And we can see that previous state-of-the-art in LDAM--DRW improved those two minority classes to 0.63 and 0.66. However, our proposed MAMix--DRW further elevated the per class accuracy of C8 and C9 and \textbf{0.79} and \textbf{0.82} respectively, without sacrificing the performance of the majority classes, which can be another evidence that shows the effectiveness of our algorithm.

Moreover, as seen in Table~\ref{tab10}, in the proposed MAMix, we can
simply set $\omega$ to $0.25$, which is consistent with that suggested for 
LDAM~\cite{CITEKC2019}; however, the performance changes little when using
different settings for $\omega$, demonstrating that the proposed method is 
easy to tune.

Furthermore, we also compared our method with two additional methods: (1)~Major-to-minor translation (M2m)~\cite{M2mCVPR2020}. (2)~Fusing deep learning and SMOTE for imbalance data (DeepSMOTE)~\cite{DeepSMOTE2021} to have a variety of comparisons. The results in Tables~\ref{tab7}, ~\ref{tab8}, ~\ref{tab9}, ~\ref{tab11}, ~\ref{tab12} reveal that our proposed method not only performs better on all five datasets but also requires less training time and computational cost. To be more precise, M2m spent much time translating the majority sample to the minority by adding noise into the images and then using the pre-trained model to label them. Meanwhile, DeepSMOTE requires more time and server computing capacity because this method has two stages. The first stage is to train the DeepSMOTE model, and the second stage uses this model to generate synthetic data. Generally, both M2m and DeepSMOTE are much more complicated to implement in comparison to our proposed method.

\subsection{Margin Perspectives}

\subsubsection{Proposed Metric: Margin Gap}
To better analyze and quantify the effect of different learning algorithms on
the majority- and minority-class margins, we define the 
margin gap metric $\gamma_{\mathit{gap}}$ as
\begin{equation}
    \gamma_{\mathit{gap}} = \dfrac{\sum_{i} n_{i} \cdot \overline{\gamma_{i}}}{\sum_{i} n_{i}} - 
    \dfrac{\sum_{j} n_{j} \cdot \overline{\gamma_{j}}}{\sum_{j} n_{j}}, \label{eq9}
\end{equation}
where $i$, $j$ belong to majority and minority classes,
respectively. To decide which class belongs to a majority class, and which belongs
a minority class, we set a threshold: if the class sample numbers
exceed ${1} \mathbin{/} {K}$ of the total training samples, we categorize them
as majority classes; the others are viewed as minority classes.

Hence a large margin gap corresponds to majority classes with larger margins   
and minority classes with smaller margins, and hence poor generalizability for the minority classes. We
hope to achieve a smaller margin gap when given unbalanced classes.
Note that this metric can be negative, as the margins for minority
classes are larger than those of majority classes. To better determine whether this is a
good indicator of the correlation between the margin gap
and top-1 validation accuracy, we further evaluate with Spearman's rank order
correlation $\rho$ in Fig.~\ref{fig2}.

\subsubsection{Spearman's Rank Order Correlation}
\begin{figure}
    \centering
    \includegraphics[width=.6\linewidth]{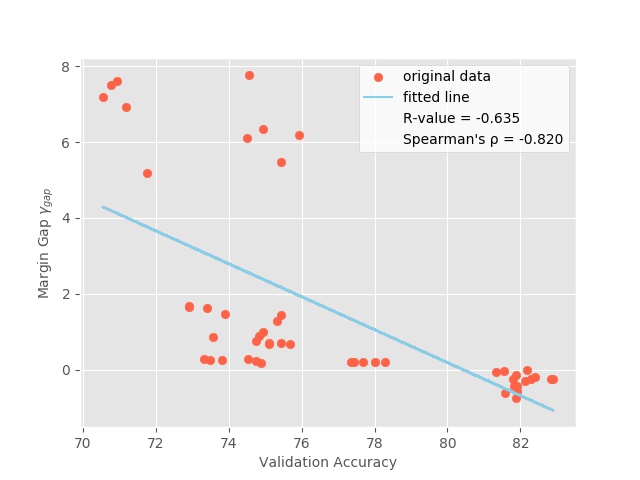}
	 \caption{Relationship between margin gap and validation accuracy for long-tailed 
	 imbalanced CIFAR-10 with imbalance ratio $\rho$ = 100 using
	 ResNet32}
    \label{fig2}
\end{figure}

We demonstrate the results of analysis using Spearman's rank order
correlation in Fig.~\ref{fig2}. We note a negative rank order correlation between
validation accuracy and margin gap $\gamma_{\mathit{gap}}$, 
as our definition of margin gap reflects the 
trend in which the better the model generalizes to the minority class, the lower
the margin gap is. That is, better models produce smaller margin gaps between
majority and minority classes. As seen in Fig.~\ref{fig2}, 
Spearman's rank order correlation is -0.820, showing that although it is sometimes
noisy, in general $\gamma_{\mathit{gap}}$ is a good indicator for
top-1 validation accuracy. Note that we will discuss the noisy part later in the next subsection.

\subsubsection{Uneven Margin}
Given the superior empirical performance of Mixup-based methods, we further analyzed
this from a margin perspective to demonstrate the effectiveness of our method.
First, we establish our baseline margin gap when the model is
trained using ERM. Then, we examine the margin-based LDAM work in which
larger margins are enforced for minority classes~\cite{CITEKC2019}. As seen
in Table~\ref{tab4}, the margin gap for ERM is the highest; that
is, for deep models trained using ERM, majority classes tend to have higher
margins than minority classes, resulting in poor
generalizability for minority classes. LDAM-DRW~\cite{CITEKC2019}
demonstrates its ability to shrink the margin gap, reducing the generalization
error for the minority class through margin-based softmax training. Moreover, we
observe that in long-tailed imbalance, the original Mixup
alone yields competitive results, as the margin gaps are
similar between the original Mixup, Remix, and our proposed
method. This observation is consistent with Remix, for which
similar performance is reported in a long-tailed imbalance setting. However,
in a step imbalance setting, the superiority of our method 
is evident, as it not only achieves better performance but also shrinks
the margin gap more than the original Mixup. 

Note that in Table~\ref{tab4}, we see that for the long-tailed scenario, the margin gap
of Remix-DRW is -1.598 and that of MAMix-DRW is -1.136. However, 
as shown in Table~\ref{tab7},
their respective validation accuracies are 81.82 and 82.29. This is an example of the noisy part that is mentioned in the previous context. Here Remix-DRW yields a smaller margin gap
than that of MAMix-DRW but poorer validation accuracy, 
because Remix tends to enforce excessive margins in minority classes, whereas our method
strikes a better trade-off.

To further study why excessive margins in minority classes do not help with validation accuracy, we first decompose the margins into two parts: $\gamma \geq 0$ and $\gamma < 0$ part, where validation accuracy is decided by the $\gamma < 0$ part ($\gamma < 0$ determines the validation error). The detailed decomposition result is in Table~\ref{tabexcessive}, where we take all $\gamma < 0$ margins and report the average among majority classes and minority classes for each method, and we compute $\gamma \geq 0$ part the same way. From our observation, $\gamma < 0$ part is generally similar between Remix and our MAMix, thus there is only slight accuracy difference, however, the $\gamma \geq 0$ part is generally higher for Remix, as we can see from Table~\ref{tabexcessive}. Therefore, the reason why in this case Remix has lower margin gap lies in the fact that it enforces more margins in $\gamma \geq 0$ part of minority classes, as we can see the $\gamma \geq 0$ part is 4.891 for Remix minority classes, and 4.213 for that of MAMix counterpart. From this observation, we identify that there seems to be \emph{excessive margins} in minority classes for Remix, but---Do these excessive margins help or not ?---Previous research~\cite{CITERL2006} has indicated that overly optimizing the margin may be an over-kill, in which the performance may be worse. We further answer this question by examining the difference between theoretical and practical margin distribution.

\begin{table}
\caption{Margin decomposition on long-tailed
imbalanced CIFAR-10 with $\rho=100$ using ResNet32 (Majority: Class 0 to Class 2; Minority: Class 3 to Class 9)}
\begin{center}
\begin{tabular}{@{}cccccccccccc@{}}
\toprule
{Average Margin} & $\gamma < 0$ &  $\gamma \geq 0$  \\ 
\midrule
{Remix--DRW Majority} & -1.587 & 2.371 \\
{MAMix--DRW Majority} & -1.523 & 2.308 \\
{Remix--DRW Minority} & -1.933 & \textbf{4.891} \\
{MAMix--DRW Minority} & -1.875 & \textbf{4.213} \\

\bottomrule
\end{tabular}
\label{tabexcessive}
\end{center}
\end{table}

Recall that LDAM~\cite{CITEKC2019} derives a theoretically optimal ratio \eqref{eq1} for per class margin distribution, where such a ratio hints the need to \emph{not over-push} the margin of minority classes. To further analyze how close the practical per class margin distribution of different methods are than that of theoretical margin distribution, we fit theoretical margin by practical margin, and since there is a constant multiplier $C$ in theoretical margin, as in the form of~\eqref{eq1}, we choose to use linear regression without bias. We set $C=1$ and compare the fitting ($L_2$) error in Table~\ref{tabL2error}. As we can see from Table~\ref{tabL2error}, our proposed MAMix shows the smallest $L_2$ error, hinting that the per class margin distribution produced by our method is the \emph{closest} to the theoretical margin distribution derived by~\cite{CITEKC2019}, while the per class margin distribution produced by Remix~\cite{CITEHC2020} is slightly inferior than ours in terms of $L_2$ error between theoretical and practical margin, which is due to the excessive margins in minority classes as shown in \emph{Remix-DRW Minority $\gamma \geq 0$} part in  Table~\ref{tabexcessive}. Moreover, from Table~\ref{tabL2error} and Table~\ref{tab7}, we observe that the closer practical margin is to theoretical margin, the higher the validation accuracy. Therefore, from the above evidence, we argue that we not only need to enforce larger margin for minority classes, but also need to not over-push minority margins, indicating the need for our method to strike for the better trade-off.

\begin{table}
\caption{$L_2$ error on long-tailed imbalanced CIFAR-10 with $\rho$ = 100 using ResNet32}
\begin{center}
\begin{tabular}{@{}cccccccccccc@{}}
\toprule
{Method} & $L_2$ Error \\
\midrule
ERM & 0.435 \\
LDAM--DRW & 0.195 \\
Mixup--DRW & 0.0133 \\
Remix--DRW & 0.0179 \\
MAMix--DRW & \textbf{0.0126} \\
\bottomrule
\end{tabular}
\label{tabL2error}
\end{center}
\end{table}


Note that in Table~\ref{tab5}---the extremely imbalanced setting---our method 
brings the margin gap closer than Remix, verifying that our method consistently 
outperforms Remix.

Therefore, from a margin perspective, we first establish the baseline: when
trained with ERM for imbalanced learning, the margins for majority
classes are significantly larger than those for minority classes. Second, the
recently proposed LDAM loss indeed shrinks the margin gap
significantly, suggesting that their approach is effective. To answer
the original question---Can we achieve uneven margins for class-imbalanced
learning through data augmentation?---the answer is positive, as we observe
that applying the original Mixup implicitly closes the gap from a margin
perspective, achieving comparable results. We further achieve uneven margins
explicitly through the proposed MAMix.

\section{Conclusion}
In this work, we are first to utilize margin statistics to analyze whether the
model has learned a proper representation under a class-imbalanced learning setting
from a margin perspective. We propose achieving uneven margins
via Mixup-based techniques. We first show that coupled
with DRW training, the original Mixup implicitly achieves uneven margins in general
imbalanced multi-class classification. However, in the case of extreme data
imbalance (for example, CINIC-10 with an imbalance ratio $\rho \geq$
200), the proposed margin-aware Mixup outperforms 
Mixup by explicitly controlling the degree of uneven margins, and also outperforms
the recently proposed Remix~\cite{CITEHC2020}. Therefore, in practice, we
suggest using the original Mixup for good results on general imbalanced tasks;
for extremely imbalanced tasks, we offer the proposed method to better account for
such data imbalance. In sum, our study connects 
SMOTE to Mixup in deep imbalanced classification, while shedding light on a novel
framework that combines both traditional~\cite{CITENC2002} and 
modern~\cite{CITEHC2020, CITEHZ2018} data augmentation techniques under the
same umbrella. Future work is needed to examine the theoretical aspects of
these Mixup-based approaches. With this method and our developed software, we hope that our work can serve as a starting
point for future research in the community.


\newcommand{\ra}[1]{\renewcommand{\arraystretch}{#1}}
\begin{table*}\centering
\caption{Top-1 validation accuracy (mean $\pm$ std) on imbalanced CIFAR-10
using ResNet32}
\ra{1.3}
\scalebox{0.65}{
\begin{tabular}{@{}cccccccccccc@{}}\toprule
{Dataset}& \multicolumn{3}{c}{Long-tailed} & \phantom{abc}& \multicolumn{3}{c}{Step} &
\phantom{abc}\\
\cmidrule{2-4} \cmidrule{6-8} \cmidrule{10-12}
{Imbalance ratio}& $100$ & $50$ & $10$ && $100$ & $50$ & $10$\\ \midrule
ERM & 71.23 $\pm$ 0.51 & 77.33 $\pm$ 0.74 & 86.72 $\pm$ 0.36 && 65.64 $\pm$ 0.82 & 71.41 $\pm$ 1.21 & 85.02 $\pm$ 0.33 \\
Mixup & 74.03 $\pm$ 0.96 & 78.79 $\pm$ 0.16 & 87.79 $\pm$ 0.42 && 66.91 $\pm$ 0.74 & 72.84 $\pm$ 0.60 & 85.50 $\pm$ 0.37 \\
Remix & 75.18 $\pm$ 0.26 & 80.21 $\pm$ 0.26 & 88.36 $\pm$ 0.36 && 69.26 $\pm$ 0.48 & 74.50 $\pm$ 1.16 & 86.68 $\pm$ 0.38 \\
MAMix & 74.74 $\pm$ 0.76 & 80.00 $\pm$ 0.24 & 88.17 $\pm$ 0.15 && 68.24 $\pm$ 0.43 & 73.88 $\pm$ 0.35 & 85.91 $\pm$ 0.33 \\
LDAM & 74.01 $\pm$ 0.68 & 78.71 $\pm$ 0.38 & 86.43 $\pm$ 0.32 && 65.64 $\pm$ 0.52 & 72.37 $\pm$ 0.61 & 84.74 $\pm$ 0.26 \\
DRW & 75.08 $\pm$ 0.61 & 80.11 $\pm$ 0.67 & 87.52 $\pm$ 0.25 && 72.02 $\pm$ 0.59 & 78.17 $\pm$ 0.27 & 87.73 $\pm$ 0.15 \\
M2m & 76.15 $\pm$ 0.72 & 80.71 $\pm$ 0.17 & 88.01 $\pm$ 0.24 && 72.91 $\pm$ 0.90 & 79.12 $\pm$ 0.21 & 87.85 $\pm$ 0.11 \\
DeepSMOTE & 76.66 $\pm$ 0.57 & 80.60 $\pm$ 0.38 & 87.60 $\pm$ 0.25 && 72.47 $\pm$ 0.64 & 77.52 $\pm$ 0.42 & 87.33 $\pm$ 0.07 \\
LDAM--DRW & 77.75 $\pm$ 0.39 & 81.70 $\pm$ 0.22 & 87.67 $\pm$ 0.39 && 77.99 $\pm$ 0.65 & 81.80 $\pm$ 0.39 & 87.68 $\pm$ 0.38 \\
Mixup--DRW & 82.11 $\pm$ 0.57 & 85.15 $\pm$ 0.27 & 89.28 $\pm$ 0.23 && 79.22 $\pm$ 0.98 & 83.28 $\pm$ 0.50 & 89.24 $\pm$ 0.15 \\
Remix--DRW & 81.82 $\pm$ 0.14 & 84.73 $\pm$ 0.23 & 89.33 $\pm$ 0.36 && 80.31 $\pm$ 0.70 & 83.61 $\pm$ 0.24 & 89.10 $\pm$ 0.15 \\
MAMix--DRW & \textbf{82.29 $\pm$ 0.60} & 85.11 $\pm$ 0.32 & 89.30 $\pm$ 0.14 && 80.02 $\pm$ 0.27 & 83.47 $\pm$ 0.19 & \textbf{89.29 $\pm$ 0.29} \\
\bottomrule
\end{tabular}}
\label{tab7}
\end{table*}

\begin{table*}\centering
\caption{Top-1 validation accuracy (mean $\pm$ std) on imbalanced CIFAR-100
using ResNet32}
\ra{1.3}
\scalebox{0.65}{
\begin{tabular}{@{}cccccccccccc@{}}\toprule
{Dataset}& \multicolumn{3}{c}{Long-tailed} & \phantom{abc}& \multicolumn{3}{c}{Step} &
\phantom{abc}\\
\cmidrule{2-4} \cmidrule{6-8} \cmidrule{10-12}
{Imbalance ratio}& $100$ & $50$ & $10$ && $100$ & $50$ & $10$\\ \midrule
ERM & 38.46 $\pm$ 0.36 & 43.51 $\pm$ 0.55 & 56.90 $\pm$ 0.13 && 39.56 $\pm$ 0.31 & 42.81 $\pm$ 0.21 & 55.09 $\pm$ 0.21 \\
Mixup & 40.69 $\pm$ 0.39 & 46.07 $\pm$ 0.60 & 59.63 $\pm$ 0.32 && 39.89 $\pm$ 0.10 & 41.09 $\pm$ 0.16 & 55.79 $\pm$ 0.35 \\
Remix & 42.46 $\pm$ 0.51 & 47.81 $\pm$ 0.48 & 60.71 $\pm$ 0.41 && 40.27 $\pm$ 0.18 & 42.97 $\pm$ 0.24 & 58.77 $\pm$ 0.23 \\
MAMix & 42.59 $\pm$ 0.22 & 47.89 $\pm$ 0.87 & 60.86 $\pm$ 0.55 && 40.02 $\pm$ 0.19 & 41.85 $\pm$ 0.44 & 57.39 $\pm$ 0.40 \\
LDAM & 40.49 $\pm$ 0.62 & 44.69 $\pm$ 0.37 & 56.06 $\pm$ 0.44 && 40.56 $\pm$ 0.29 & 43.11 $\pm$ 0.09 & 54.29 $\pm$ 0.41 \\
DRW & 40.40 $\pm$ 0.80 & 45.19 $\pm$ 0.49 & 57.23 $\pm$ 0.33 && 42.97 $\pm$ 0.24 & 46.78 $\pm$ 0.38 & 56.82 $\pm$ 0.38 \\
M2m & 41.92 $\pm$ 1.01 & 46.25 $\pm$ 0.15 & 58.34 $\pm$ 0.07 && 45.66 $\pm$ 0.02 & 49.54 $\pm$ 0.06 & 59.08 $\pm$ 0.22 \\
DeepSMOTE & 38.87 $\pm$ 0.19 & 44.70 $\pm$ 0.34 & 56.97 $\pm$ 0.25 && 42.27 $\pm$ 0.16 & 46.22 $\pm$ 0.39 & 55.45 $\pm$ 0.20 \\
LDAM--DRW & 41.28 $\pm$ 0.43 & 45.61 $\pm$ 0.41 & 56.42 $\pm$ 0.38 && 43.51 $\pm$ 0.61& 46.81 $\pm$ 0.29 & 56.07 $\pm$ 0.30 \\
Mixup--DRW & 46.91 $\pm$ 0.46 & 51.75 $\pm$ 0.20 & 62.18 $\pm$ 0.24 && 47.56 $\pm$ 0.34 & 53.50 $\pm$ 0.47 & 62.91 $\pm$ 0.53 \\
Remix--DRW & 46.00 $\pm$ 0.48 & 51.16 $\pm$ 0.23 & 61.63 $\pm$ 0.25 && 48.91 $\pm$ 0.29 & 53.75 $\pm$ 0.26 & 62.47 $\pm$ 0.35 \\
MAMix--DRW & \textbf{46.93 $\pm$ 0.24} & \textbf{51.92 $\pm$ 0.20} & \textbf{62.30 $\pm$ 0.33} && 48.87 $\pm$ 0.36 & \textbf{53.87 $\pm$ 0.62} & \textbf{62.84 $\pm$ 0.18} \\
\bottomrule
\end{tabular}}
\label{tab8}
\end{table*}

\begin{table*}\centering
\caption{Top-1 validation accuracy (mean $\pm$ std) on imbalanced CINIC-10
using ResNet18}
\ra{1.3}
\scalebox{0.65}{
\begin{tabular}{@{}cccccccccccc@{}}\toprule
{Dataset}& \multicolumn{3}{c}{Long-tailed} & \phantom{abc}& \multicolumn{3}{c}{Step} &
\phantom{abc}\\
\cmidrule{2-4} \cmidrule{6-8} \cmidrule{10-12}
{Imbalance ratio}& $100$ & $50$ & $10$ && $100$ & $50$ & $10$\\ \midrule
ERM & 61.08 $\pm$ 0.55 & 66.17 $\pm$ 0.37 & 77.64 $\pm$ 0.08 && 57.29 $\pm$ 0.73 & 62.26 $\pm$ 0.42 & 75.39 $\pm$ 0.30 \\
DRW & 63.75 $\pm$ 0.22 & 69.35 $\pm$ 0.35 & 78.66 $\pm$ 0.10 && 64.34 $\pm$ 0.25 & 68.73 $\pm$ 0.27 & 78.24 $\pm$ 0.21 \\
M2m & 64.20 $\pm$ 0.22 & 69.84 $\pm$ 0.41 & 78.67 $\pm$ 0.11 && 63.99 $\pm$ 1.25 & 69.82 $\pm$ 0.20 & 78.66 $\pm$ 0.03 \\
LDAM--DRW & 68.15 $\pm$ 0.22 & 72.34 $\pm$ 0.42 & 79.03 $\pm$ 0.17 && 70.09 $\pm$ 0.32 & 73.16 $\pm$ 0.48 & 79.07 $\pm$ 0.10 \\
Mixup--DRW & 71.40 $\pm$ 0.25 & 75.02 $\pm$ 0.16 & 81.36 $\pm$ 0.09 && 71.33 $\pm$ 0.23 & 74.74 $\pm$ 0.20 & 81.37 $\pm$ 0.18\\
Remix--DRW & 71.15 $\pm$ 0.24 & 74.68 $\pm$ 0.09 & 81.27 $\pm$ 0.13 && 71.48 $\pm$ 0.50 & 74.91 $\pm$ 0.21 & 81.26 $\pm$ 0.08\\
MAMix--DRW & \textbf{71.76 $\pm$ 0.29} & \textbf{75.27 $\pm$ 0.17} & \textbf{81.46 $\pm$ 0.08} && \textbf{71.91 $\pm$ 0.23} & \textbf{75.26 $\pm$ 0.08} & \textbf{81.39 $\pm$ 0.08} \\
\bottomrule
\end{tabular}}
\label{tab9}
\end{table*}

\begin{table*}\centering
\caption{Sensitivity of $\omega$ in long-tailed extremely imbalanced CIFAR-10 with
$\rho$ = 300 using ResNet32}
\begin{tabular}{@{}cccccccccccc@{}}
\toprule
{Method} & Accuracy \\
\midrule
MAMix--DRW ($\omega$ = 0.125) & 74.64 $\pm$ 0.17  \\
MAMix--DRW ($\omega$ = 0.25) & 74.85 $\pm$ 0.28 \\
MAMix--DRW ($\omega$ = 0.5) & 74.7 $\pm$ 0.75 \\
MAMix--DRW ($\omega$ = 1.0) & 74.66 $\pm$ 0.36 \\
MAMix--DRW ($\omega$ = 2.0) & 74.21 $\pm$ 0.56 \\
MAMix--DRW ($\omega$ = 4.0) & 74.05 $\pm$ 0.50 \\
MAMix--DRW ($\omega$ = 8.0) & 73.52 $\pm$ 0.52 \\
\bottomrule
\end{tabular}
\label{tab10}
\end{table*}

\begin{table*}\centering
\caption{Top-1 validation accuracy (mean $\pm$ std) on imbalanced SVHN using
ResNet32}
\ra{1.3}
\scalebox{0.65}{
\begin{tabular}{@{}cccccccccccc@{}}\toprule
{Dataset}& \multicolumn{3}{c}{Long-tailed} & \phantom{abc}& \multicolumn{3}{c}{Step} &
\phantom{abc}\\
\cmidrule{2-4} \cmidrule{6-8} \cmidrule{10-12}
{Imbalance ratio}& $100$ & $50$ & $10$ && $100$ & $50$ & $10$\\ \midrule
ERM & 79.91 $\pm$ 0.67 & 83.42 $\pm$ 0.15 & 88.43 $\pm$ 0.22 && 76.38 $\pm$ 0.93 & 81.33 $\pm$ 1.11 & 87.89 $\pm$ 0.31 \\
Mixup & 81.57 $\pm$ 0.68 & 85.16 $\pm$ 0.48 & 90.75 $\pm$ 0.28 && 76.62 $\pm$ 1.03 & 82.88 $\pm$ 1.06 & 89.79 $\pm$ 0.61 \\
Remix & 82.37 $\pm$ 0.67 & 86.27 $\pm$ 0.41 & 91.07 $\pm$ 0.21 && 78.89 $\pm$ 1.30 & 83.57 $\pm$ 0.63 & 90.20 $\pm$ 0.45 \\
Ours & 82.39 $\pm$ 0.45 & 86.75 $\pm$ 0.37 & 91.09 $\pm$ 0.25 && 77.83 $\pm$ 1.87 & 83.91 $\pm$ 0.97 & 90.68 $\pm$ 0.32 \\
LDAM & 81.96 $\pm$ 0.69 & 85.31 $\pm$ 0.29 & 89.40 $\pm$ 0.36 && 77.93 $\pm$ 1.00 & 83.84 $\pm$ 0.62 & 89.45 $\pm$ 0.37 \\
DRW & 80.68 $\pm$ 0.32 & 83.66 $\pm$ 0.49 & 88.64 $\pm$ 0.26 && 76.33 $\pm$ 2.00 & 82.29 $\pm$ 1.17 & 88.18 $\pm$ 0.45 \\
M2m & 77.68 $\pm$ 0.45 & 82.25 $\pm$ 0.36 & 88.39 $\pm$ 0.38 && 76.10 $\pm$ 0.83 & 80.46 $\pm$ 1.96 & 87.84 $\pm$ 0.77 \\
DeepSMOTE & 81.12 $\pm$ 0.58 & 83.62 $\pm$ 0.55 & 88.06 $\pm$ 0.49 && 78.67 $\pm$ 0.88 & 82.08 $\pm$ 0.52 & 87.73 $\pm$ 0.19 \\
LDAM--DRW & 83.48 $\pm$ 1.11 & 86.17 $\pm$ 0.54 & 89.85 $\pm$ 0.26 && 79.24 $\pm$ 1.19 & 84.79 $\pm$ 0.65 & 90.11 $\pm$ 0.41\\
Mixup--DRW & 85.19 $\pm$ 0.32 & 87.43 $\pm$ 0.63 & 90.14 $\pm$ 0.23 && 80.73 $\pm$ 1.72 & 87.32 $\pm$ 0.87 & 90.84 $\pm$ 0.24\\
Remix--DRW & 84.52 $\pm$ 0.62 & 87.27 $\pm$ 0.37 & 90.11 $\pm$ 0.53 && 80.90 $\pm$ 1.96 & 87.09 $\pm$ 0.85& 90.80 $\pm$ 0.23\\
MAMix--DRW & \textbf{85.41 $\pm$ 0.56} & \textbf{87.79 $\pm$ 0.45} & \textbf{90.59 $\pm$ 0.52} && \textbf{81.71 $\pm$ 1.28} & \textbf{87.62 $\pm$ 0.36} & 90.57 $\pm$ 0.23 \\
\bottomrule
\end{tabular}}
\label{tab11}
\end{table*}

\begin{table*}\centering
\caption{Top-1 validation accuracy (mean $\pm$ std) on imbalanced Tiny-ImageNet
using ResNet18}
\ra{1.3}
\scalebox{0.65}{
\begin{tabular}{@{}cccccccccccc@{}}\toprule
{Dataset}& \multicolumn{2}{c}{Long-tailed} & \phantom{ab} &  \multicolumn{2}{c}{Step} & \phantom{ab} \\
\cmidrule{2-3} \cmidrule{5-6}
{Imbalance ratio}& $100$ & $10$ && $100$  & $10$\\ \midrule
{ERM} & 32.86 $\pm$ 0.22 & 48.90 $\pm$ 0.43 && 35.44 $\pm$ 0.25 & 48.23 $\pm$ 0.13 \\
{DRW} & 33.81 $\pm$ 0.49 & 49.99 $\pm$ 0.27 && 37.79 $\pm$ 0.11 & 50.13 $\pm$ 0.30 \\
{M2m} & 34.33 $\pm$ 0.42 & 49.39 $\pm$ 0.63 && 37.02 $\pm$ 0.68 & 50.11 $\pm$ 0.24 \\
{LDAM} & 31.13 $\pm$ 0.36 & 46.90 $\pm$ 0.19 && 35.88 $\pm$ 0.09 &  47.91 $\pm$ 0.19 \\
{LDAM--DRW} & 31.90 $\pm$ 0.13 & 47.15 $\pm$ 0.31 && 36.75 $\pm$ 0.19 & 48.17 $\pm$ 0.16 \\
{Mixup--DRW} & 37.97 $\pm$ 0.38 & 52.51 $\pm$ 0.40 && 40.45 $\pm$ 0.21 & 54.46 $\pm$ 0.29 \\
{Remix--DRW} & 36.89 $\pm$ 0.61 & 52.13 $\pm$ 0.23 && 41.07 $\pm$ 0.37 & 53.58 $\pm$ 0.23 \\
{MAMix--DRW} & 37.73 $\pm$ 0.18 & 52.53 $\pm$ 0.34 && \textbf{41.46 $\pm$ 0.38} & 54.37 $\pm$ 0.29 \\
\bottomrule
\end{tabular}}
\label{tab12}
\end{table*}

\begin{table*}\centering\tiny
\caption{Per Class Accuracy in long-tailed imbalanced CIFAR-10 with
$\rho$ = 100 using ResNet32}
\begin{tabular}{ c c c c c c c c c c c c c c }
\toprule
{Method} & C0 & C1 & C2 & C3 & C4 & C5 & C6 & C7 & C8 & C9  \\
\midrule
ERM & 0.94 & 0.97 & 0.83 & 0.71 & 0.76 & 0.61 & 0.72 & 0.61 & 0.46 & 0.48  \\
LDAM--DRW  & 0.95 & 0.97 & 0.79 & 0.73 & 0.82 & 0.69 & 0.78 & 0.70 & 0.63 & 0.66  \\
MAMix--DRW  & 0.89 & 0.94 & 0.79 & 0.71 & 0.82 & 0.76 & 0.85 & \textbf{0.81} & \textbf{0.79} & \textbf{0.82} \\
\bottomrule
\end{tabular}
\label{tab15}
\end{table*}

\begin{appendices}

\section{Implementation Details}

\subsection{Implementation Details for CIFAR}
We followed \cite{CITEKC2019} for CIFAR-10 and CIFAR-100. We also followed \cite{CITEKC2019} to perform simple data augmentation described in \cite{CITEHE2016} for training, where we first padded 4 pixels on each side, then a 32 x 32 crop was randomly sampled from the padded image, or its horizontal flip. We also used ResNet-32 \cite{CITEHE2016} as our base network. We trained the model with a batch size of 128 for 200 epochs. We use an initial learning rate of 0.1, then decay by 0.01 at the 160 and 180th epoch. We also use linear warm-up learning rate schedule \cite{CITEGP2018} for the first 5
epochs for fair comparison.

\subsection{Implementation Details fo CINIC}
We followed \cite{CITEHC2020} for CINIC-10 where we used ResNet-18 \cite{CITEHE2016} as our base network. Initially, the training scheme provided by \cite{CITEHC2020} was to train the model for 300 epochs, with initial learning rate of 0.1 and decay the learning rate by 0.01 at the 150th, and 225th epoch. However, we found that training for 200 epochs is sufficient, thus we trained the model for 200 epochs, with a batch size of 128, and initial learning rate of 0.1, followed by decaying the learning rate by 0.01 at the 160 and 180th epochs. We also use linear warm-up learning rate schedule. When DRW was deployed, it was deployed at the 225th epoch. When LDAM was used, we enforced the largest margin to be 0.5. 

\subsection{Implementation Details for SVHN}
We followed \cite{CITEYY2020} for SVHN. We adoped ResNet-32 \cite{CITEHE2016} as our base network. We trained the model for 200 epochs, with initial learning rate of 0.1 and batch size of 128. We used linear warm-up schedule, and decay the learning rate by 0.1 at the 160th, and 180th epochs. When DRW was deployed, it was deployed at the 160th epoch. When LDAM was used, we enforced the largest margin to be 0.5.

\subsection{Implementation Details for Tiny ImageNet}
We followed \cite{CITEKC2019} for Tiny ImageNet with 200 classes. For basic data augmentation in training, we first performed simple horizontal flips, followed by taking random crops of size 64 x 64 from images padded by 8 pixels on each side. We adopted ResNet-18 \cite{CITEHE2016} as our base networks, and used stochastic gradient descent with momentum of 0.9, weight decay of $2 \cdot 10^{-4}$. We trained the model for 120 epochs, with initial learning rate of 0.1 and batch size of 128. We used linear warm-up rate schedule, and decay the learning rate by 0.1 at the 90th epoch. When DRW was deployed, it was deployed at the 90th epoch. When LDAM was used, we follow the original paper to enforce largest margin to be 0.5. Note that we cannot reproduce the numbers reported in \cite{CITEKC2019}.




\end{appendices}


\section*{Declarations}
\begin{itemize}
\item Funding: The work is mainly supported by the MOST of Taiwan under 107-2628-E-002-008-MY3.
 \item Conflicts of interest/Competing interests: n/a
 \item Ethics approval: n/a
 \item Consent to participate: n/a
 \item Consent for publication: n/a
 \item Availability of data and material: experiments are based on public benchmark data
 \item Code availability: released at open-source at \url{https://github.com/ntucllab/imbalanced-DL}
\item Authors' contributions: Cheng contributes to detailed literature survey, the initial idea of studying Mixup for deep imbalanced classification, experimental comparison, code implementation and release and initial manuscript writing; Ha rigorously reviewed the code implementation, addressed issues and bugs, and expanded the scope of experimental comparison by incorporating additional methods like DeepSMOTE and M2m; Lin contributes to the bigger picture of linking SMOTE and Mixup, the initial idea of the margin-aware extension, and suggestions on the research methodology.

\end{itemize}

\bibliography{sn-bibliography}


\end{document}